\documentclass[11pt,a4paper,dvipsnames]{article}
\usepackage[hyperref]{acl2021}
\usepackage[T1]{fontenc}
\usepackage[utf8]{inputenc}

\usepackage{times}
\usepackage{latexsym}
\usepackage{booktabs}
\usepackage{microtype}
\usepackage{todonotes}
\usepackage{enumitem}
\usepackage{graphicx}
\usepackage{xspace}
\usepackage{amsmath}
\usepackage{multirow}
\usepackage{multicol}
\usepackage{caption}
\usepackage{subcaption}
\usepackage{subfiles}
\usepackage{colortbl}
\usepackage{framed}
\usepackage{xcolor}
\usepackage{amsfonts}
\usepackage{makecell}
\usepackage{bbm}
\usepackage{pifont}

\newcommand{\tanda}{\textsc{TandA}\xspace}
\newcommand{\gpd}{\text{WQA}\xspace}

\newcommand{\GenQA}{GenQA\xspace}
\newcommand{\MSNLG}{MSNLG\xspace}
\newcommand{\UQAT}{\textsc{Uqa}T5\xspace}

\newcommand{\newcameraready}[1]{\textcolor{black}{#1}}

\newcommand{\colorA}[1]{\textcolor[HTML]{e79f00}{#1}\xspace}
\newcommand{\colorB}[1]{\textcolor[HTML]{999999}{#1}\xspace}
\newcommand{\colorC}[1]{\textcolor[HTML]{009e73}{#1}\xspace}
\newcommand{\colorD}[1]{\textcolor[HTML]{0171b2}{#1}\xspace}
\newcommand{\colorE}[1]{\textcolor[HTML]{cb79a7}{#1}\xspace}

\newcommand{\colorHuman}[1]{\textcolor[HTML]{2d59a6}{#1}\xspace}
\newcommand{\colorAuto}[1]{\textcolor[HTML]{740324}{#1}\xspace}

\DeclareMathOperator{\selector}{\mathcal{S}}
\DeclareMathOperator{\generator}{\mathcal{G}}

\title{Answer Generation for Retrieval-based Question Answering Systems}

\aclfinalcopy %
\def\aclpaperid{3401} %

\author{
  Chao-Chun Hsu\textsuperscript{1}\thanks{\hspace{1em}This work was completed while the author was an intern at Amazon Alexa.}\hspace{.3em}, Eric Lind\textsuperscript{2}, Luca Soldaini\textsuperscript{2}, Alessandro Moschitti\textsuperscript{2} \\
  \textsuperscript{1}University of Chicago, \textsuperscript{2}Amazon Alexa \\
  \texttt{chaochunh@uchicago.edu},  \texttt{\{lssoldai,ericlind,amosch\}@amazon.com} \\
}

\begin{document}
\maketitle
\begin{abstract}
Recent advancements in transformer-based models have greatly improved the ability of Question Answering (QA) systems to provide correct answers; 
in particular, answer sentence selection (AS2) models, core components of retrieval-based systems, have achieved impressive results.
While generally effective, these models fail to provide a satisfying answer when all retrieved candidates are of poor quality, even if they contain correct information.
In AS2, models are trained to select the best answer sentence among a set of candidates retrieved for a given question.
In this work, we propose to generate answers from a set of AS2 top candidates.
Rather than selecting the best candidate, we train a sequence to sequence transformer model to generate an answer from a candidate set. 
Our tests on three English AS2 datasets show improvement up to 32 absolute points in accuracy over the state of the art.
\end{abstract}

\section{Introduction}

Question answering systems are a core component of many commercial applications, ranging from task-based dialog systems to general purpose virtual assistants, e.g., Google Home, Amazon Alexa, and Siri.
Among the many approaches for QA, AS2 has attracted significant attention in the last few years  \citep{Tymoshenko2018CrossPairTR,Tian2020CapturingSR,garg2020tanda,han-etal-2021-modeling}.
Under this framework, for a given question, a retrieval system is first used to obtain and rank a set of supporting passages;
then, an AS2 model is used to estimate the likelihood of each sentence extracted from passages to be a correct answer, returning the one with the highest probability.
This approach is favored in virtual assistant systems because full sentences are more likely to include the right context and sound natural, both of which are characteristics users value \cite{berdasco2019ux}.

AS2 models have shown great performance on academic benchmarks.
However, these datasets fail to consider many essential qualities of a QA system which interacts directly with users, such as a virtual assistant.
In some cases, extracted answer sentences contain the correct information, but the focus of the answer doesn't match the question;
in others, the answer requires reasoning or contextual knowledge from the user or is very long and contains extraneous information. 
For example, in WikiQA \cite{yang2015wikiqa}, a widely used AS2 dataset, the answer
\textit{``Wind power is the conversion of wind energy into a useful form of energy, such as using wind turbines to make electrical power,  windmills for mechanical power, wind pumps for water pumping...
''} 
is considered a good answer for \textit{``What can be powered by wind?''}, even though its formulation is burdensome to a user.

\setlength{\tabcolsep}{2pt}
\begin{table}[t]
\footnotesize
\centering
\renewcommand{\arraystretch}{1.2}
\resizebox{\linewidth}{!}{%
\begin{tabular}{lp{7.2cm}}
\toprule
$\mathbf{Q}$: & {How a \textbf{\colorA{water}} \textbf{\colorB{pump}} works?}\\
$\mathbf{c_1}$: & A small, electrically powered \textbf{\colorB{pump}}.\\
$\mathbf{c_2}$: & A large, electrically driven \textbf{\colorB{pump}} (electropump) for \textbf{\colorA{waterworks}} near the Hengsteysee,  Germany.\\
$\mathbf{c_3}$: & A \textbf{\colorB{pump}} \textbf{\colorC{is a device}} that \textbf{\colorD{moves fluids}} (liquids or gases), or sometimes slurries, by \textbf{\colorE{mechanical action}}.\\
$\mathbf{c_4}$: & \textbf{\colorB{Pumps}} can be classified into three major groups according to the method they use to \textbf{\colorD{move the fluid}}: direct lift, displacement, and gravity \textbf{\colorB{pumps}}.\\
$\mathbf{c_5}$: & \textbf{\colorB{Pumps}} operate by some mechanism (typically reciprocating or rotary), and consume energy to perform \textbf{\colorE{mechanical work}} by \textbf{\colorD{moving the fluid}}.\\
\midrule
$\mathbf{G}$: & A \textbf{\colorA{water}} \textbf{\colorB{pump}} \textbf{\colorC{is a device}} that \textbf{\colorD{moves fluids}} by \textbf{\colorE{mechanical action}}.\\
\bottomrule
\end{tabular}
}
\caption{An example of a question Q and five answer candidates $c_1,\ldots,c_5$ from WikiQA \cite{yang2015wikiqa} ranked by an AS2 system. 
Answer $G$ generated by our best system is significantly more natural and concise than any extracted candidates.}
\label{ex:intro}
\end{table}

In this work, we explore a fundamentally different approach to AS2.
Rather than \textit{selecting} the best candidate, we propose using a  model to \textit{generate} a suitable response for a user question. 
\newcameraready{In so doing, we extend the traditional AS2 pipeline with a final generation stage that can recover correct and satisfying answers in cases where a ranking AS2 model fails to place an acceptable candidate at the top position or where a top candidate with the desired information is not a natural-sounding response to the query.}
Table~\ref{ex:intro} shows an example of our system: given the question, $Q$, and a list of candidates, $C_k= \left\{c_1, \ldots, c_5\right\}$ sorted by a state-of-the-art AS2 system, we use a sequence-to-sequence model to produce an answer $G$ given $Q$ and $C_k$ as input.
This approach, which we refer to as \GenQA, addresses the limitations of AS2 systems by composing concise answers which may contain information from multiple sources.

Recent works have shown that large, transformer-based conditional generative models can be used to significantly improve parsing \cite{chen-etal-2020-low,Rongali2020DontPG},
retrieval \cite{de2020autoregressive,pradeep2021expando}, and classification tasks \cite{DBLP:journals/corr/abs-1910-10683}.
Our approach builds on top of this line of work by designing and testing generative models for AS2-based QA systems.
In recent years, the use of generative approaches has been evaluated for other QA tasks, such as machine reading (MR) \cite{izacard2020leveraging,lewis2020retrieval} and question-based summarization (QS) \cite{iida2019exploiting,goodwin-etal-2020-towards,deng2020joint}. 
However, while related, these efforts are fundamentally different from the experimental setting described in this paper.
Given a question, generative MR models are used to extract a short span (1-5 tokens) from a passage that could be used to construct an answer to a question.  In contrast, AS2 returns a complete sentence that could be directly returned to a user. 

QS systems are designed to create a general summary given a question and one or more related documents. 
Unlike QS, AS2-based QA systems need to provide specific answers; thus, the presence of even a small amount of unrelated information in a response could cause the answer sentence to be unsuitable. 
In contrast, we show that our approach can succinctly generate the correct information from a set of  highly relevant sentence candidates.

In summary, our contribution is four-fold: (\textit{i}) we introduce a new approach for AS2-based QA systems, which generates, rather than selects, an answer sentence; 
(\textit{ii}) we illustrate how to adapt state-of-the-art models such as T5 \cite{DBLP:journals/corr/abs-1910-10683} and BART \cite{lewis2019bart} for answer generation;  (\textit{iii}) we show\footnote{Our models, source code, and  annotated data are available at: \texttt{\url{https://github.com/alexa/wqa-cascade-transformers}}.} that our \GenQA system improves over the state-of-the-art AS2-based systems by up to 32 accuracy points, as evaluated by human annotators;
finally, (\textit{iv}) we briefly explain why traditional generation metrics are not suited for evaluating AS2-based systems.

\section{Datasets}
We use four English datasets in our work, one related to generative QA and three to AS2. 
For a fair comparison between selector and generation methods, we re-evaluate the top answers returned by all models using a fixed set of annotators.
All annotations were completed by company associates who are not part of our research group and had no knowledge of the systems.
Annotators were required to mark an answer as correct if it was: (\textit{i}) factually correct; (\textit{ii}) natural-sounding; and (\textit{iii}) required no additional information to be understood.
All QA pairs were single annotated, as we determined sufficient agreement for this task in previous campaigns.

\paragraph{WikiQA}\hspace{-1em} by \citet{yang2015wikiqa} contains queries from Bing search logs and candidate answer sentences extracted from a relevant Wikipedia page. 
For evaluation, we used the dev.~and test sets, which contain 126 and 243 unique questions and 
we re-annotated all of the resulting 569 QA pairs.\footnote{\newcameraready{Due to time and annotation constraints, we were only able to annotate results for 100 queries from each of the dev.~and test sets for our \UQAT model}}

\paragraph{Answer Sentence Natural Questions (ASNQ)} introduced by \citet{garg2020tanda} was derived from the NQ dataset \cite{kwiatkowski2019natural} and consists of the questions which have a short answer span within a single sentence in a long answer span.  The sentences containing the short answer are marked as correct and the other sentences in the document are marked as incorrect.  
We use the dev.~and test splits introduced by \citet{soldaini-moschitti-2020-cascade} 
which contain 1,336 questions each.
We re-annotated a total of 5,344 QA pairs.

\paragraph{\gpd} \hspace{-1em} is an internal AS2 dataset created from a non-representative sample of questions asked by users of a virtual personal assistant in 2019\footnote{\newcameraready{The public version of \gpd will be released in the short-term future. Please search for a publication with title \emph{WQA: A Dataset for Web-based Question Answering Tasks} on arXiv.}}. 
For each question, we retrieved 500 pages from an index containing over 100M web documents.
We then ranked candidate answers using a state-of-the-art AS2 system, and annotated up to 100 of them. 
In total, the training and dev.~sets contain 3,074 queries and 189k QA pairs, while the test set contains 808 queries.
For this effort, we re-annotated 4,847 QA pairs from the test set.

\paragraph{MS MARCO QA NLG (MSNLG)} \hspace{-1em} by \citet{nguyen2016ms} is a subset of the MS MARCO dataset focused on generating natural language answers to user queries from web search result passages. 
It consists of 182k queries from Bing search logs, the ten most relevant passages retrieved for each query, and a well-formed answer synthesized by an annotator.
This dataset is not designed for AS2, but it represents a large resource of succinct and clear answers, thus making it close to our AS2 task.

\section{Generative QA Model (\GenQA)}

The AS2 task is defined as follows:
Let $q$ be an element of the question set, $Q$, and $C_q=\{c_1, \dots, c_n\}$ be a set of candidates for $q$,  e.g., sentences retrieved by a search engine, where $c_i \in C$,  and $C$ is a set of candidates. 
We model a selector $\selector: Q \times C^n \rightarrow C$, such that $\selector(q,C_q) = \text{argmax}_i \left(p(q,c_i)\right)$, where $p(q,c_i)$ is the probability that $c_i$ is a good answer.
We also define $\selector_k: Q \times C^n \rightarrow C^k$, such that, $\selector_k$ selects the top $k$ answers in descending order of $p(q,c_i)$.

\paragraph{State of the Art} Throughout our experiments, we use \tanda \citep{garg2020tanda} as our state-of-the-art selector $\selector$. This AS2 model was trained as a binary classifier on $(q,c_i)$ pairs using a sequential fine-tuning approach starting with ASNQ and finishing on a target dataset, e.g., WikiQA.  
Specifically, we use their pretrained RoBERTa Large model  \cite{liu2019roberta}, as it achieved the best results on all datasets it was tested on.

\subsection{Our Generative Approach}
Instead of selecting the best candidate, we generate a new answer using the information from the top $k$ answer candidates. Thus, our model is a function $\generator:Q \times C^k \rightarrow G$, where $G$ is the text that can be generated by the generator $\generator$ from the question, any fragment of the retrieval set, the model's vocabulary, and knowledge stored in the model's parameters.  Formally: 
\begin{equation}
\generator(q,C_q)=\generator(q, C_k)=\generator(q, \selector_k(k, {C_q})).
\label{Gen}
\end{equation}

The example in Table~\ref{ex:intro} shows that we can generate a correct answer from a set of candidates which, as a whole, contain enough information to formulate a correct answer. 
We propose that a valid answer can be built by composing the \emph{most promising} constituents coming from the different candidates in $C_k$. 
Intuitively, information repeated across multiple candidates is more \emph{promising}; 
therefore, we hypothesize that a model trained on the same or similar generation task should be able to effectively exploit this form of repetition, even in cases where the same information is presented in a similar, but not identical manner. 
Further, recent works have shown that large transformer models hold a substantial amount of commonsense knowledge in their parameters \cite{roberts-etal-2020-much}, which our model could leverage to perform inference across sentences in $C_k$, \textit{e.g.}, associate \textit{water} with \textit{fluid} in the example in Table~\ref{ex:intro}.

\subsection{Fine-tuning \GenQA}
\label{tuningS}
Given a pre-trained transformer seq2seq model, e.g., T5 \cite{DBLP:journals/corr/abs-1910-10683} or BART \cite{lewis2019bart}, we obtain $\generator$ by fine-tuning on a large AS2 or QA generation dataset.
For this purpose, we format our training data as a standard sequence-to-sequence/text-to-text task, where the source text is the question concatenated with the top five answer candidates, $(q,\selector_{k=5}$), joined by newlines.  
When an answer composed by a human is available, such as in \MSNLG, we use it as the output target. 
For cases where there is no composed answer, we randomly select a known-good candidate to be the target, remove it from the inputs and replace it with another candidate if one is available.  
We truncate the input text to 512 tokens and, at test time, we use beam search with a beam size of four and a maximum output length of 100 tokens.

\section{Experiments}
In this section, we first report on our experimental setup, then we show the results on fine-tuning GenQA, and finally, we report on the comparative results between AS2 and GenQA.
\subsection{Setup}
\label{setup}

\paragraph{Models and Parameterization}
Our \GenQA model is based on the T5 \cite{DBLP:journals/corr/abs-1910-10683} variant of the UnifiedQA (\UQAT) model by \citet{UnifiedQA2020}. 
We use the Large version of \UQAT, which has 770M parameters for all of our experiments.  
We compute training loss as the mean of the cross-entropy between the softmax probabilities over the output vocabulary and the one-hot encoded target answer.  
We fine-tune \UQAT with a learning rate of $5E^{-5}$.
We also experiment with the Large variant of BART \cite{lewis2019bart}, which is comprised of 400M parameters.
This model was trained using same loss with learning rate $5E^{-6}$.

\paragraph{Evaluation}
We used accuracy as our primary metric for all our experiments and models. This is computed as the average number of questions a model answers correctly;
for a selector $\selector$, it is equivalent to Precision at 1.
For $\selector$, we also report Hit Rate at 5, which is the fraction of queries with at least one good candidate ranked five or less. 

Beside human evaluation, we also experimented with automatic evaluation metrics such as BLEU \cite{papineni2002bleu} and ROUGE-L \cite{lin2004rouge} for \GenQA. 
Such metrics have found little success in evaluating QA tasks \cite{chaganty-etal-2018-price,chen-etal-2019-evaluating}, so we investigate whether that is the case for AS2 as well.

\subsection{Results}

\paragraph{How to Fine-tune \GenQA?}
As described in Section~\ref{setup}, we tested two \GenQA variants: 
one uses a UnifiedQA T5 (\UQAT) \cite{UnifiedQA2020} as base model, while the other leverages BART-Large \cite{lewis2019bart}. 
Of the datasets used in this work, \MSNLG and \gpd are large enough for fine-tuning \GenQA. 
Therefore, based on preliminary results, we tested four different strategies for training \UQAT: 
fine tuning on (\textit{i}) \gpd or (\textit{ii}) \MSNLG alone, (\textit{iii}) combine the two datasets by alternating mini-batches during training, or (\textit{iv}) follow the \textit{transfer-then-adapt} strategy proposed by \citet{garg2020tanda}: first fine-tune on \MSNLG, then adapt to a AS2 using \gpd.

\begin{table}[t]
\centering
\renewcommand{\arraystretch}{1.2}
\small
\begin{tabular}{lc@{\hspace{8pt}}cc}
\toprule
\textbf{Model} & \colorHuman{\footnotesize\textbf{Accuracy}} & \colorAuto{\textsc{\textbf{Bleu}}} & \colorAuto{\textsc{\textbf{Rouge-l}}} \\
\midrule
\tanda \cite{garg2020tanda} & \colorHuman{\textit{baseline}} & \colorAuto{-} & \colorAuto{-} \\
\midrule
\UQAT (AS2D) &                  \colorHuman{+5.3\%} & \colorAuto{40.8} & \colorAuto{55.7} \\
\UQAT (\MSNLG) &                \colorHuman{\textbf{+19.9\%}} & \colorAuto{20.2} & \colorAuto{39.7} \\
\UQAT ({\MSNLG}+AS2D) &           \colorHuman{+13.6\%} & \colorAuto{35.3} & \colorAuto{50.6} \\
\UQAT ({\MSNLG}$\rightarrow$AS2D) & \colorHuman{+7.9\%} & \colorAuto{40.6} & \colorAuto{54.8} \\
\midrule
BART-Large (\MSNLG) & \colorHuman{\textbf{+20.7\%}} & \colorAuto{21.5} & \colorAuto{41.1}  \\
\bottomrule
    \end{tabular}
    \caption{
    Relative accuracy of different \GenQA models and training configurations on the \gpd dataset;
    both \UQAT and BART perform best when finetuned on \MSNLG only. 
    As shown in previous work, \colorAuto{automatic metrics (\textsc{Bleu}, \textsc{Rouge-l})} do not correlate with \colorHuman{human annotations (accuracy)}.
    }
    \label{tab:results2}
\end{table}

Table~\ref{tab:results2} reports the results on the \gpd test set, which  are all relative to the performance of the state-of-the-art model (\tanda). 
First, we observe that all \GenQA models reported in this table considerably outperform the best selector model, \tanda. 
This result shows that our generative approach can improve system based on AS2.

Comparing the accuracy of different training strategies applied to \UQAT, we achieve the best results when the model is trained on \MSNLG alone (+19.9\% over \tanda baseline). 
While we were initially surprised by this result, as \MSNLG is not designed for AS2, error analysis suggests that \GenQA benefits from the high quality training data (concise answers written by annotators). 
Conversely, when training with \gpd, we observed that \GenQA tends to produce answers that, while correct, are not as natural-sounding. 
We plan to explore how to best leverage existing AS2 datasets for generative model training in future work. We also note that a \GenQA BART-Large achieves comparable results to \GenQA \UQAT on \gpd; in preliminary experiments, we found training strategies reported on \UQAT to have similar effect on BART-Large.  

\newcameraready{When manually annotating results of our early tests, we found that BART was more likely to be extractive and copy input passages in their entirety while \UQAT was more likely to compose new text and produce answers with textual overlap from multiple input candidates but was more likely to hallucinate content.  We found that through hyperparameter tuning we could largely eliminate the hallucination from \UQAT answers but we were unable to make BART more abstractive. 
}

Similar to what has been observed in other QA tasks \cite{chaganty-etal-2018-price,chen-etal-2019-evaluating}, we find that automatic metrics do not correlate with assessments from human annotators. 
This is due to the fact that neither BLEU nor ROUGE-L are designed to estimate whether an answer is clear and natural-sounding, instead rewarding candidates that have high overlap with reference answers. Most importantly, such overlap is a poor indicator of factual correctness.

\paragraph{Comparison between AS2 and \GenQA}
Table~\ref{tab:results1} reports the results of \tanda and \GenQA on two standard AS2 datasets, evaluated with manual annotation. We note that there is an impressive gap of over 20 absolute accuracy points on both development and test sets.
This result is produced by two important properties of \GenQA. 
First, it builds correct answers from a pool of correct and incorrect answers, and it can generate a good answer so long as the relevant information can be found anywhere in the top $k=5$ candidates. 
This is a clear advantage over using \tanda alone, as Hit-Rate@5 of 99.2\%, and 87.9\% for WikiQA and ASNQ, respectively, ensures that \GenQA often receives at least one correct answer as input.

\begin{table}[t]
\small
\renewcommand{\arraystretch}{1.2}
\centering
    \begin{tabular}{lccc@{\hspace{8pt}}cc}
        \toprule
        & \multicolumn{3}{c}{\textbf{\tanda}} & \multicolumn{2}{c}{\textbf{\GenQA \UQAT}} \\
        \textbf{Dataset} & \textbf{Acc.} & \textbf{Hit@5} & \textbf{Length}    & \textbf{Acc.} & \textbf{Length}  \\
        \midrule
        WikiQA\textsubscript{DEV}     & 59.5  & 99.2 & $31.7\pm13.7$     & \textbf{92.1}  & $14.9\pm9.3$\\
        WikiQA\textsubscript{TEST}    & 61.0  & 99.2 & $30.1\pm12.4$     & \textbf{88.5}  & $14.6\pm8.3$\\
        \midrule
        ASNQ\textsubscript{DEV}       & 75.5  & 87.7 & $41.0\pm122.4$   & \textbf{90.2} & $13.9\pm5.9$\\
        ASNQ\textsubscript{TEST}      & 69.0  & 87.9 & $37.9\pm51.5$    & \textbf{90.5} & $13.9\pm5.6$\\
        \bottomrule
    \end{tabular}
\caption{Accuracy of our \GenQA \UQAT model compared to a state-of-the-art AS2 model by \citet{garg2020tanda}.  
All answer candidates returned by the two models were re-annotated to ensure a fair comparison.  
Length is the average number of tokens in the answer.}
\label{tab:results1}
\end{table}

Second, \GenQA exhibits the ability to rewrite \emph{unnatural} answers from a text snippet into an answer suitable for a conversation. 
For example, for the question \textit{``What year did Isaac Newton die?''},
\tanda returns candidate \textit{``Sir Isaac Newton (25 December 1642--20 March 1727) was an English physicist and mathematician''}. Although correct, no human would provide it in such a form. 
In contrast, \GenQA composes a concise answer: \emph{``Isaac Newton died in 1727''}.

Finally, Table~\ref{tab:results1} shows that the size of \GenQA answers, in terms of words, is only 14 tokens, which is 2.7 times less than the 30-40 tokens from \tanda. This further suggests that \GenQA can provide more concise and direct answers, which are preferable in a conversational context.

\section{Conclusions}
In this work we present \GenQA, a generative approach for AS2-based QA systems. 
\newcameraready{The main difference with recent MR-based generative systems is the capacity of our models to generate long answers. This comes from the use of AS2 candidates (complete sentences) as input to our generative approach. In contrast, MR systems, being mainly trained with short answers, e.g., noun phrases and named entities, mostly generate short answers.}

We show that \GenQA significantly outperforms state-of-the-art selector models for AS2 by up to 32 accuracy points by combining different pieces of information from the top $k$ answer candidates.
These results suggest promising directions for generative retrieval-based systems.

\section*{Acknowledgments}

\newcameraready{We thank Thuy Vu for setting up annotation procedures for the \gpd dataset.}

\bibliography{custom}

\begin{thebibliography}{27}
\expandafter\ifx\csname natexlab\endcsname\relax\def\natexlab#1{#1}\fi

\bibitem[{Berdasco et~al.(2019)Berdasco, López, Diaz, Quesada, and
  Guerrero}]{berdasco2019ux}
Ana Berdasco, Gustavo López, Ignacio Diaz, Luis Quesada, and Luis~A. Guerrero.
  2019.
\newblock User experience comparison of intelligent personal assistants: Alexa,
  google assistant, siri and cortana.
\newblock \emph{Proceedings}, 31(1).

\bibitem[{Chaganty et~al.(2018)Chaganty, Mussmann, and
  Liang}]{chaganty-etal-2018-price}
Arun Chaganty, Stephen Mussmann, and Percy Liang. 2018.
\newblock The price of debiasing automatic metrics in natural language
  evalaution.
\newblock In \emph{Proceedings of the 56th Annual Meeting of the Association
  for Computational Linguistics (Volume 1: Long Papers)}, pages 643--653,
  Melbourne, Australia. Association for Computational Linguistics.

\bibitem[{Chen et~al.(2019)Chen, Stanovsky, Singh, and
  Gardner}]{chen-etal-2019-evaluating}
Anthony Chen, Gabriel Stanovsky, Sameer Singh, and Matt Gardner. 2019.
\newblock Evaluating question answering evaluation.
\newblock In \emph{Proceedings of the 2nd Workshop on Machine Reading for
  Question Answering}, pages 119--124, Hong Kong, China. Association for
  Computational Linguistics.

\bibitem[{Chen et~al.(2020)Chen, Ghoshal, Mehdad, Zettlemoyer, and
  Gupta}]{chen-etal-2020-low}
Xilun Chen, Asish Ghoshal, Yashar Mehdad, Luke Zettlemoyer, and Sonal Gupta.
  2020.
\newblock Low-resource domain adaptation for compositional task-oriented
  semantic parsing.
\newblock In \emph{Proceedings of the 2020 Conference on Empirical Methods in
  Natural Language Processing (EMNLP)}, pages 5090--5100, Online. Association
  for Computational Linguistics.

\bibitem[{De~Cao et~al.(2020)De~Cao, Izacard, Riedel, and
  Petroni}]{de2020autoregressive}
Nicola De~Cao, Gautier Izacard, Sebastian Riedel, and Fabio Petroni. 2020.
\newblock Autoregressive entity retrieval.
\newblock \emph{arXiv preprint arXiv:2010.00904}.

\bibitem[{Deng et~al.(2020)Deng, Lam, Xie, Chen, Li, Yang, and
  Shen}]{deng2020joint}
Yang Deng, Wai Lam, Yuexiang Xie, Daoyuan Chen, Yaliang Li, Min Yang, and Ying
  Shen. 2020.
\newblock Joint learning of answer selection and answer summary generation in
  community question answering.
\newblock In \emph{The Thirty-Fourth {AAAI} Conference on Artificial
  Intelligence, {AAAI} 2020, The Thirty-Second Innovative Applications of
  Artificial Intelligence Conference, {IAAI} 2020, The Tenth {AAAI} Symposium
  on Educational Advances in Artificial Intelligence, {EAAI} 2020, New York,
  NY, USA, February 7-12, 2020}, pages 7651--7658. {AAAI} Press.

\bibitem[{Garg et~al.(2020)Garg, Vu, and Moschitti}]{garg2020tanda}
Siddhant Garg, Thuy Vu, and Alessandro Moschitti. 2020.
\newblock {TANDA:} transfer and adapt pre-trained transformer models for answer
  sentence selection.
\newblock In \emph{The Thirty-Fourth {AAAI} Conference on Artificial
  Intelligence, {AAAI} 2020, The Thirty-Second Innovative Applications of
  Artificial Intelligence Conference, {IAAI} 2020, The Tenth {AAAI} Symposium
  on Educational Advances in Artificial Intelligence, {EAAI} 2020, New York,
  NY, USA, February 7-12, 2020}, pages 7780--7788. {AAAI} Press.

\bibitem[{Goodwin et~al.(2020)Goodwin, Savery, and
  Demner-Fushman}]{goodwin-etal-2020-towards}
Travis Goodwin, Max Savery, and Dina Demner-Fushman. 2020.
\newblock Towards {Z}ero-{S}hot {C}onditional {S}ummarization with {A}daptive
  {M}ulti-{T}ask {F}ine-{T}uning.
\newblock In \emph{Findings of the Association for Computational Linguistics:
  EMNLP 2020}, pages 3215--3226, Online. Association for Computational
  Linguistics.

\bibitem[{Han et~al.(2021)Han, Soldaini, and
  Moschitti}]{han-etal-2021-modeling}
Rujun Han, Luca Soldaini, and Alessandro Moschitti. 2021.
\newblock Modeling context in answer sentence selection systems on a latency
  budget.
\newblock In \emph{Proceedings of the 16th Conference of the European Chapter
  of the Association for Computational Linguistics: Main Volume}, pages
  3005--3010, Online. Association for Computational Linguistics.

\bibitem[{Iida et~al.(2019)Iida, Kruengkrai, Ishida, Torisawa, Oh, and
  Kloetzer}]{iida2019exploiting}
Ryu Iida, Canasai Kruengkrai, Ryo Ishida, Kentaro Torisawa, Jong{-}Hoon Oh, and
  Julien Kloetzer. 2019.
\newblock Exploiting background knowledge in compact answer generation for
  why-questions.
\newblock In \emph{The Thirty-Third {AAAI} Conference on Artificial
  Intelligence, {AAAI} 2019, The Thirty-First Innovative Applications of
  Artificial Intelligence Conference, {IAAI} 2019, The Ninth {AAAI} Symposium
  on Educational Advances in Artificial Intelligence, {EAAI} 2019, Honolulu,
  Hawaii, USA, January 27 - February 1, 2019}, pages 142--151. {AAAI} Press.

\bibitem[{Izacard and Grave(2021)}]{izacard2020leveraging}
Gautier Izacard and Edouard Grave. 2021.
\newblock Leveraging passage retrieval with generative models for open domain
  question answering.
\newblock In \emph{Proceedings of the 16th Conference of the European Chapter
  of the Association for Computational Linguistics: Main Volume}, pages
  874--880, Online. Association for Computational Linguistics.

\bibitem[{Khashabi et~al.(2020)Khashabi, Min, Khot, Sabharwal, Tafjord, Clark,
  and Hajishirzi}]{UnifiedQA2020}
Daniel Khashabi, Sewon Min, Tushar Khot, Ashish Sabharwal, Oyvind Tafjord,
  Peter Clark, and Hannaneh Hajishirzi. 2020.
\newblock {UNIFIEDQA}: Crossing format boundaries with a single {QA} system.
\newblock In \emph{Findings of the Association for Computational Linguistics:
  EMNLP 2020}, pages 1896--1907, Online. Association for Computational
  Linguistics.

\bibitem[{Kwiatkowski et~al.(2019)Kwiatkowski, Palomaki, Redfield, Collins,
  Parikh, Alberti, Epstein, Polosukhin, Devlin, Lee, Toutanova, Jones, Kelcey,
  Chang, Dai, Uszkoreit, Le, and Petrov}]{kwiatkowski2019natural}
Tom Kwiatkowski, Jennimaria Palomaki, Olivia Redfield, Michael Collins, Ankur
  Parikh, Chris Alberti, Danielle Epstein, Illia Polosukhin, Jacob Devlin,
  Kenton Lee, Kristina Toutanova, Llion Jones, Matthew Kelcey, Ming-Wei Chang,
  Andrew~M. Dai, Jakob Uszkoreit, Quoc Le, and Slav Petrov. 2019.
\newblock Natural questions: A benchmark for question answering research.
\newblock \emph{Transactions of the Association for Computational Linguistics},
  7:452--466.

\bibitem[{Lewis et~al.(2020{\natexlab{a}})Lewis, Liu, Goyal, Ghazvininejad,
  Mohamed, Levy, Stoyanov, and Zettlemoyer}]{lewis2019bart}
Mike Lewis, Yinhan Liu, Naman Goyal, Marjan Ghazvininejad, Abdelrahman Mohamed,
  Omer Levy, Veselin Stoyanov, and Luke Zettlemoyer. 2020{\natexlab{a}}.
\newblock {BART}: Denoising sequence-to-sequence pre-training for natural
  language generation, translation, and comprehension.
\newblock In \emph{Proceedings of the 58th Annual Meeting of the Association
  for Computational Linguistics}, pages 7871--7880, Online. Association for
  Computational Linguistics.

\bibitem[{Lewis et~al.(2020{\natexlab{b}})Lewis, Perez, Piktus, Petroni,
  Karpukhin, Goyal, K{\"{u}}ttler, Lewis, Yih, Rockt{\"{a}}schel, Riedel, and
  Kiela}]{lewis2020retrieval}
Patrick S.~H. Lewis, Ethan Perez, Aleksandra Piktus, Fabio Petroni, Vladimir
  Karpukhin, Naman Goyal, Heinrich K{\"{u}}ttler, Mike Lewis, Wen{-}tau Yih,
  Tim Rockt{\"{a}}schel, Sebastian Riedel, and Douwe Kiela. 2020{\natexlab{b}}.
\newblock Retrieval-augmented generation for knowledge-intensive {NLP} tasks.
\newblock In \emph{Advances in Neural Information Processing Systems 33: Annual
  Conference on Neural Information Processing Systems 2020, NeurIPS 2020,
  December 6-12, 2020, virtual}.

\bibitem[{Lin(2004)}]{lin2004rouge}
Chin-Yew Lin. 2004.
\newblock {ROUGE}: A package for automatic evaluation of summaries.
\newblock In \emph{Text Summarization Branches Out}, pages 74--81, Barcelona,
  Spain. Association for Computational Linguistics.

\bibitem[{Liu et~al.(2019)Liu, Ott, Goyal, Du, Joshi, Chen, Levy, Lewis,
  Zettlemoyer, and Stoyanov}]{liu2019roberta}
Yinhan Liu, Myle Ott, Naman Goyal, Jingfei Du, Mandar Joshi, Danqi Chen, Omer
  Levy, Mike Lewis, Luke Zettlemoyer, and Veselin Stoyanov. 2019.
\newblock Roberta: A robustly optimized bert pretraining approach.
\newblock \emph{arXiv preprint arXiv:1907.11692}.

\bibitem[{Nguyen et~al.(2016)Nguyen, Rosenberg, Song, Gao, Tiwary, Majumder,
  and Deng}]{nguyen2016ms}
Tri Nguyen, Mir Rosenberg, Xia Song, Jianfeng Gao, Saurabh Tiwary, Rangan
  Majumder, and Li~Deng. 2016.
\newblock Ms marco: A human generated machine reading comprehension dataset.

\bibitem[{Papineni et~al.(2002)Papineni, Roukos, Ward, and
  Zhu}]{papineni2002bleu}
Kishore Papineni, Salim Roukos, Todd Ward, and Wei-Jing Zhu. 2002.
\newblock {B}leu: a method for automatic evaluation of machine translation.
\newblock In \emph{Proceedings of the 40th Annual Meeting of the Association
  for Computational Linguistics}, pages 311--318, Philadelphia, Pennsylvania,
  USA. Association for Computational Linguistics.

\bibitem[{Pradeep et~al.(2021)Pradeep, Nogueira, and Lin}]{pradeep2021expando}
Ronak Pradeep, Rodrigo Nogueira, and Jimmy Lin. 2021.
\newblock The expando-mono-duo design pattern for text ranking with pretrained
  sequence-to-sequence models.
\newblock \emph{arXiv preprint arXiv:2101.05667}.

\bibitem[{Raffel et~al.(2019)Raffel, Shazeer, Roberts, Lee, Narang, Matena,
  Zhou, Li, and Liu}]{DBLP:journals/corr/abs-1910-10683}
Colin Raffel, Noam Shazeer, Adam Roberts, Katherine Lee, Sharan Narang, Michael
  Matena, Yanqi Zhou, Wei Li, and Peter~J. Liu. 2019.
\newblock Exploring the limits of transfer learning with a unified text-to-text
  transformer.
\newblock \emph{CoRR}, abs/1910.10683.

\bibitem[{Roberts et~al.(2020)Roberts, Raffel, and
  Shazeer}]{roberts-etal-2020-much}
Adam Roberts, Colin Raffel, and Noam Shazeer. 2020.
\newblock How much knowledge can you pack into the parameters of a language
  model?
\newblock In \emph{Proceedings of the 2020 Conference on Empirical Methods in
  Natural Language Processing (EMNLP)}, pages 5418--5426, Online. Association
  for Computational Linguistics.

\bibitem[{Rongali et~al.(2020)Rongali, Soldaini, Monti, and
  Hamza}]{Rongali2020DontPG}
Subendhu Rongali, Luca Soldaini, Emilio Monti, and Wael Hamza. 2020.
\newblock Don't parse, generate! {A} sequence to sequence architecture for
  task-oriented semantic parsing.
\newblock In \emph{{WWW} '20: The Web Conference 2020, Taipei, Taiwan, April
  20-24, 2020}, pages 2962--2968. {ACM} / {IW3C2}.

\bibitem[{Soldaini and Moschitti(2020)}]{soldaini-moschitti-2020-cascade}
Luca Soldaini and Alessandro Moschitti. 2020.
\newblock The cascade transformer: an application for efficient answer sentence
  selection.
\newblock In \emph{Proceedings of the 58th Annual Meeting of the Association
  for Computational Linguistics}, pages 5697--5708, Online. Association for
  Computational Linguistics.

\bibitem[{Tian et~al.(2020)Tian, Zhang, Feng, Jiang, Lyu, Liu, and
  Zhao}]{Tian2020CapturingSR}
Zhixing Tian, Yuanzhe Zhang, Xinwei Feng, Wenbin Jiang, Yajuan Lyu, K.~Liu, and
  Jun Zhao. 2020.
\newblock Capturing sentence relations for answer sentence selection with
  multi-perspective graph encoding.
\newblock In \emph{AAAI}.

\bibitem[{Tymoshenko and Moschitti(2018)}]{Tymoshenko2018CrossPairTR}
Kateryna Tymoshenko and Alessandro Moschitti. 2018.
\newblock Cross-pair text representations for answer sentence selection.
\newblock In \emph{Proceedings of the 2018 Conference on Empirical Methods in
  Natural Language Processing}, pages 2162--2173, Brussels, Belgium.
  Association for Computational Linguistics.

\bibitem[{Yang et~al.(2015)Yang, Yih, and Meek}]{yang2015wikiqa}
Yi~Yang, Wen-tau Yih, and Christopher Meek. 2015.
\newblock {W}iki{QA}: A challenge dataset for open-domain question answering.
\newblock In \emph{Proceedings of the 2015 Conference on Empirical Methods in
  Natural Language Processing}, pages 2013--2018, Lisbon, Portugal. Association
  for Computational Linguistics.

\end{thebibliography}
\bibliographystyle{acl_natbib}

\end{document}